\documentclass[11pt,a4paper]{article}
\usepackage[hyperref]{emnlp2020}
\usepackage{times}
\usepackage{latexsym}
\usepackage{algorithm}
\usepackage{algorithmic}
\usepackage{multirow}
\usepackage{amsmath}
\usepackage{amsfonts,amssymb}
\usepackage{subfigure}
\usepackage{graphicx}
\usepackage{comment}
\usepackage{balance}

\usepackage{microtype}

\aclfinalcopy 

\title{Wasserstein Distance Regularized Sequence Representation \\for Text Matching in Asymmetrical Domains}

\author{Weijie Yu$^1$, Chen Xu$^2$, Jun Xu$^{2,3,*}$, Liang Pang$^{4}$,
Xiaopeng Gao$^5$,\\\textbf{ Xiaozhao Wang$^5$, Ji-Rong Wen$^{2,3}$}\\

$^1$School of Information, Renmin University of China \\
$^2$Gaoling School of Artificial Intelligence, Renmin University of China\\
$^3$Beijing Key Laboratory of Big Data Management and Analysis Methods\\
$^4$Institute of Computing Technology, Chinese Academy of Sciences; $^5$Alibaba Group\\

{ \{yuweijie, xc\_chen, junxu, jrwen\}@ruc.edu.cn}, pangliang@ict.ac.cn, 
  }

\date{}

\begin{document}
\maketitle
\begin{abstract}
One approach to matching texts from asymmetrical domains is projecting the input sequences into a common semantic space as feature vectors upon which the matching function can be readily defined and learned. In real-world matching practices, it is often observed that with the training goes on, the feature vectors projected from different domains tend to be indistinguishable. The phenomenon, however, is often overlooked in existing matching models. As a result, the feature vectors are constructed without any regularization, which inevitably increases the difficulty of learning the downstream matching functions. In this paper, we propose a novel match method tailored for text matching in asymmetrical domains, called WD-Match. In WD-Match, a Wasserstein distance-based regularizer is defined to regularize the features vectors projected from different domains. As a result, the method enforces the feature projection function to generate vectors such that those correspond to different domains cannot be easily discriminated. The training process of WD-Match amounts to a game that minimizes the matching loss regularized by the Wasserstein distance. WD-Match can be used to improve different text matching methods, by using the method as its underlying matching model. Four popular text matching methods have been exploited in the paper. Experimental results based on four publicly available benchmarks showed that WD-Match consistently outperformed the underlying methods and the baselines.\let\thefootnote\relax\footnotetext{$^*$ Corresponding author}

\end{abstract}

\section{Introduction}
Asymmetrical text matching, which predicts the relationship (e.g., category, similarity) of two text sequences from different domains, is a fundamental problem in both information retrieval (IR) and natural language processing (NLP).
For example, in natural language inference (NLI), text matching has been used to determine whether a hypothesis is entailment, contradiction, or neutral given a premise~\cite{SNLI}. 
In question answering (QA), text matching has been used to determine whether a answer can answer the given question~\cite{TRECQA,wikiqa}.
In IR, text matching has been widely used to measure the relevance of a document to a query~\cite{Li:FnTIR2014:SMSearch, Xu:FNTIR:DeepMatch}. 

\begin{figure*}

\centering
\subfigure[Epoch 1]{
\centering
\includegraphics[width=0.33\linewidth]{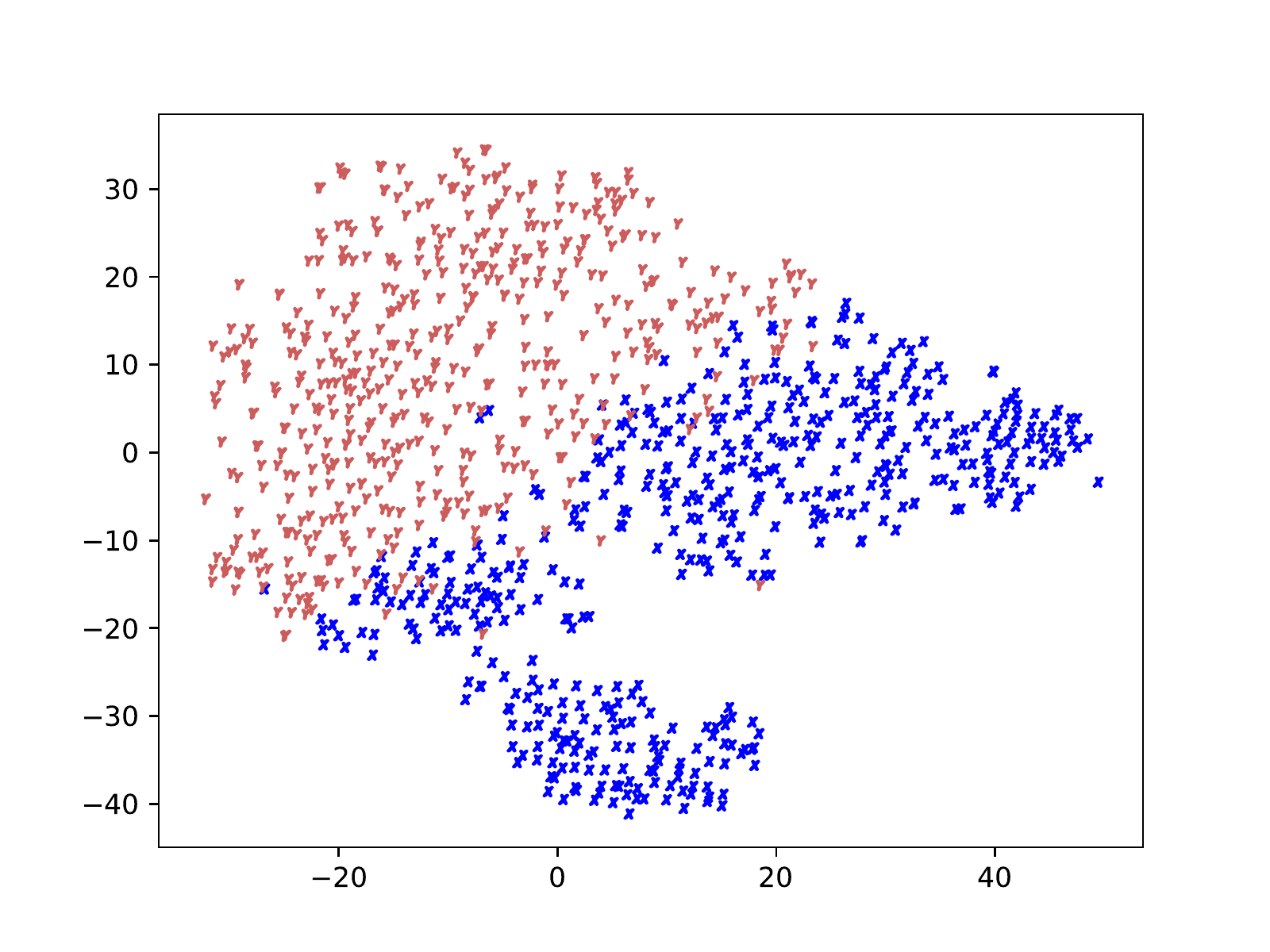}
}%
\subfigure[Epoch 10]{
\centering
\includegraphics[width=0.33\linewidth]{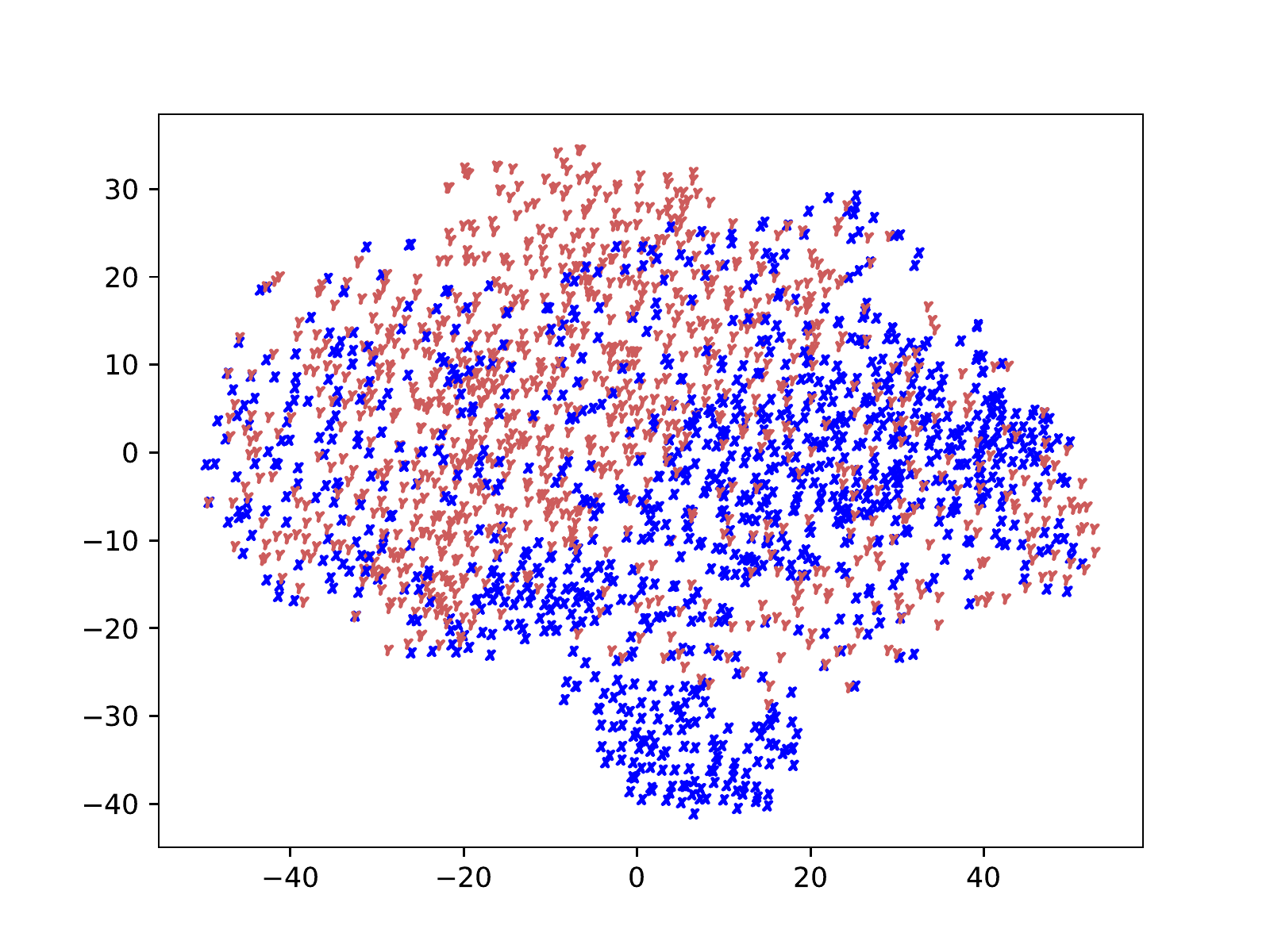}
}%
\subfigure[Epoch 20]{
\centering
\includegraphics[width=0.33\linewidth]{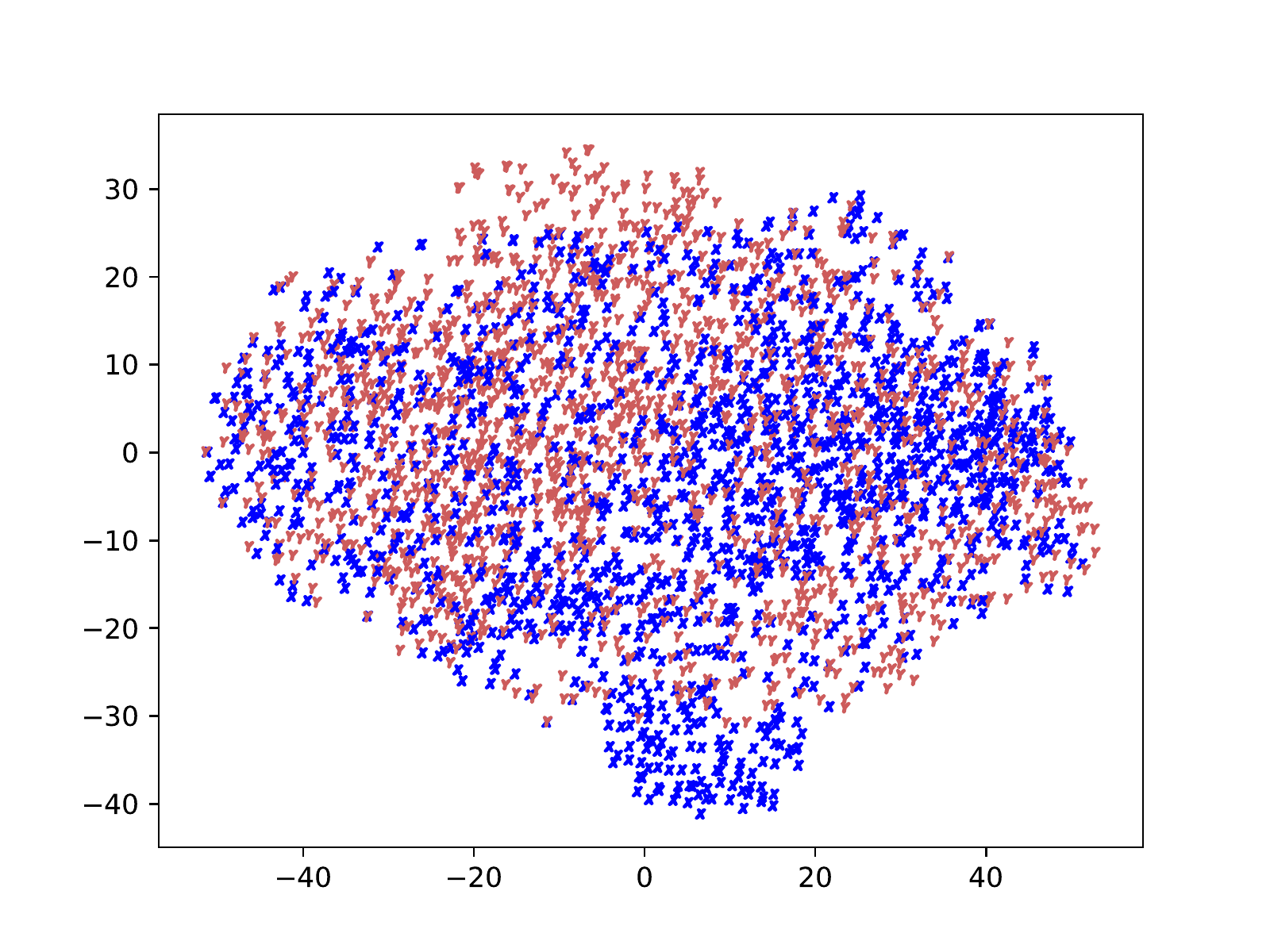}
}%
\caption{t-SNE visualization of the projected feature vectors, based on the RE2 models trained on SciTail dataset. Subfigure (a), (b), and (c) respectively illustrates the vector distributions at epochs 1, 10, and 20. The blue `X' and red `Y' correspond to the premise and the hypothesis respectively.}\label{fig:VisualizeRE2}
\label{fig:iter}
\end{figure*}

One approach to asymmetrical text matching is projecting the text sequences from different domains into a common latent space as feature vectors. Since these feature vectors have identical dimensions and in the same space, matching functions can be readily defined and learned. 
This type of approach includes a number of popular methods, such as DSSM~\cite{DSSM}, DecAtt~\cite{DecAtt}, CAFE~\cite{CAFE}, and RE2~\cite{RE2}.
In real-world matching practices, it is often observed that learning of the matching models is a process of moving the projected feature vectors together in the semantic space. For example, Figure~\ref{fig:VisualizeRE2} shows the distribution of the feature vectors generated by RE2. During the training of RE2~\cite{RE2} on SciTail dataset~\cite{Scitail}, it is observed that at the early stage of the training, the feature vectors corresponding to different domains are often separately distributed (according to the visualization by tNSE~\cite{t-SNE}) ( Figure~\ref{fig:VisualizeRE2}(a)). With the training went on, these separated feature vectors gradually moved closer and finally mixed together ( Figure~\ref{fig:VisualizeRE2}(b) and (c)). 

The phenomenon can be explained as follows. Given two text sequences from two asymmetrical domains (e.g., NLI), the first sequence (e.g., premise) and the second sequence (e.g., hypothesis) are heterogeneous and there exists a lexical gap that needs to be bridged between them~\cite{HCRN}, similar to that of learning cross-modal matching model~\cite{ACMR}. Existing studies~\cite{ACMR,cross-modal3} have shown that it is essentially critical that the projection network should generate domain- or modal-invariant features. That is, the global distributions of feature vectors should be similar in a common subspace such that their origins cannot be discriminated. The phenomenon is not unique but recurs in the experiments based on other matching models and other datasets.


Existing text matching models, however, are still lack of constraints or regularizations to ensure that the projected vectors are well distributed for matching. One natural question is, can we design a mechanism that can explicitly guide the mix of the feature vectors and better distribute them. To answer the question, this paper presents a novel learning to match method in which the Wasserstein distance (between the two distributions respectively corresponding to the two asymmetrical domains) is introduced as a regularizer, called WD-Match. WD-Match consists of three components: (1) a feature projection component which jointly projects each pair of text sequences into a latent semantic space, as a pair of feature vectors; (2) a regularizer component which estimates the Wasserstein distance with a feed-forward neural network on the basis of the projected features; (3) a matching component which conducts the matching, also on the same set of projected features.

The training of WD-Match amounts to repeatedly interplays between two branches under the adversarial learning framework: a regularizer branch that learns a neural network for estimating the upper bound on the dual form Wasserstein distance, and a matching branch that minimizes a Wasserstein distance regularized matching loss. In this way, the minimization of the loss function leads to a learning method not only to minimize the matching loss, but also to well distribute the feature vectors in the semantic space for better matching. 

To summarize, this paper makes the following main contributions: 
\begin{itemize}
    \item We highlight the critical importance of the global distribution of the projected feature vectors in matching texts between asymmetrical domains, which has not yet been seriously studied in existing models. 
    \item We propose a new learning to match method under the adversarial framework, in which the text matching model is learned by minimizing a Wasserstein distance-regularized matching loss. 
    \item We conducted empirical studies on four large scale benchmarks, and demonstrated that WD-Match achieved better performance than the baselines and the underlying models. Extensive analysis showed the effects of Wasserstein distance-based regularizer in terms of guiding the distributions of feature vectors and improving the matching accuracy. 
\end{itemize}
The source code of WD-Match is available at \url{https://github.com/RUC-WSM/WD-Match}
\section{Related Work}
In this section, we first review the sequence representation used in text matching, then introduce the Wasserstein distance and its applications.

\subsection{\mbox{Sequence Representation in Text Matching}}
Sequence representation lies in the core of text matching~\cite{Xu:FNTIR:DeepMatch}.
Early works inspired by Siamese architecture assign respective neural networks to encode two input sequences into high-level representations. For example, DSSM~\cite{DSSM} is one of the classic representation-based matching approaches to text matching which uses feed-forward neural networks to project a text sequence. CDSSM~\cite{CDSSM}, ARC-I~\cite{ARC-I} and CNTN~\cite{CNTN} change sequence encoder to a convolutional neural network which shares parameters in a fixed size sliding window. To further capture the long-term dependence of a text sequence, a group of recurrent neural network based methods were proposed, including RNN-LSTM~\cite{RNN} and MV-LSTM~\cite{MV-LSTM}.

Recently, with the help of attention mechanism~\cite{DecAtt}, the sequence representation is obtained by aligning the sequence itself and the other sequence in the input pairs.
For example, CSRAN~\cite{CSRAN"} performs multi-level attention refinement with dense
connections among multiple levels. DRCN~\cite{DRCN} stacks encoding layers and attention layers, then concatenates all previously aligned results. RE2~\cite{RE2} introduces a consecutive architecture
based on augmented residual connection between convolutional layers and attention layers. These models yield strong performance on several benchmarks.

\subsection{Wasserstein Distance}
Wasserstein distance~\cite{w_distance} is a metric based on the theory of optimal transport. It gives a natural measure of the distance between two probability distributions.

Wasserstein distance has been successfully used in the Generative Adversarial Networks (GAN)~\cite{GAN} framework of deep learning. \citet{WGAN} propose WGAN which uses the Wasserstein-1 metric as a way to improve the original framework of GAN, to alleviate the vanishing gradient and the mode collapse issues in the original GAN. The Wasserstein distance has also been explored to learn the domain-invariant features in domain adaptation tasks. For example, ~\citet{w_distance} propose to minimize the Wasserstein distance between the feature distributions of the source and the target domains, yielding better performance and smoother training than the standard training method with a Gradient Reversal Layer~\cite{DA}. ~\citet{domainadapt} propose to learn domain-invariant features with the guidance of Wasserstein distance. 

Inspired by its success in variant applications, this paper introduces Wasserstien distance to text matching in asymmetrical domains, as a regularizer to improve the sequence representations.

\section{Our Approach: WD-Match}
In this section, we describe our proposed method WD-Match.
   

\subsection{Model Architecture}\label{sec:model}
Suppose that we are given a collection of $N$ instances of sequence-sequence-label triples: $\mathcal{D} = \{(X_i, Y_i, \mathbf{z}_i)\}_{i=1}^N$ where $X_i\in \mathcal{X}$, $Y_i \in \mathcal{Y}$, and $\mathbf{z}_i\in \mathcal{Z}$ respectively denote the first sequence, the second sequence, and the label indicating the relationship of $X_i$ and $Y_i$.
As shown in Figure~\ref{fig:WD-Match_arch}, WD-Match consists of three components:
\begin{figure*}[ht]
\centering
\includegraphics[width=\textwidth]{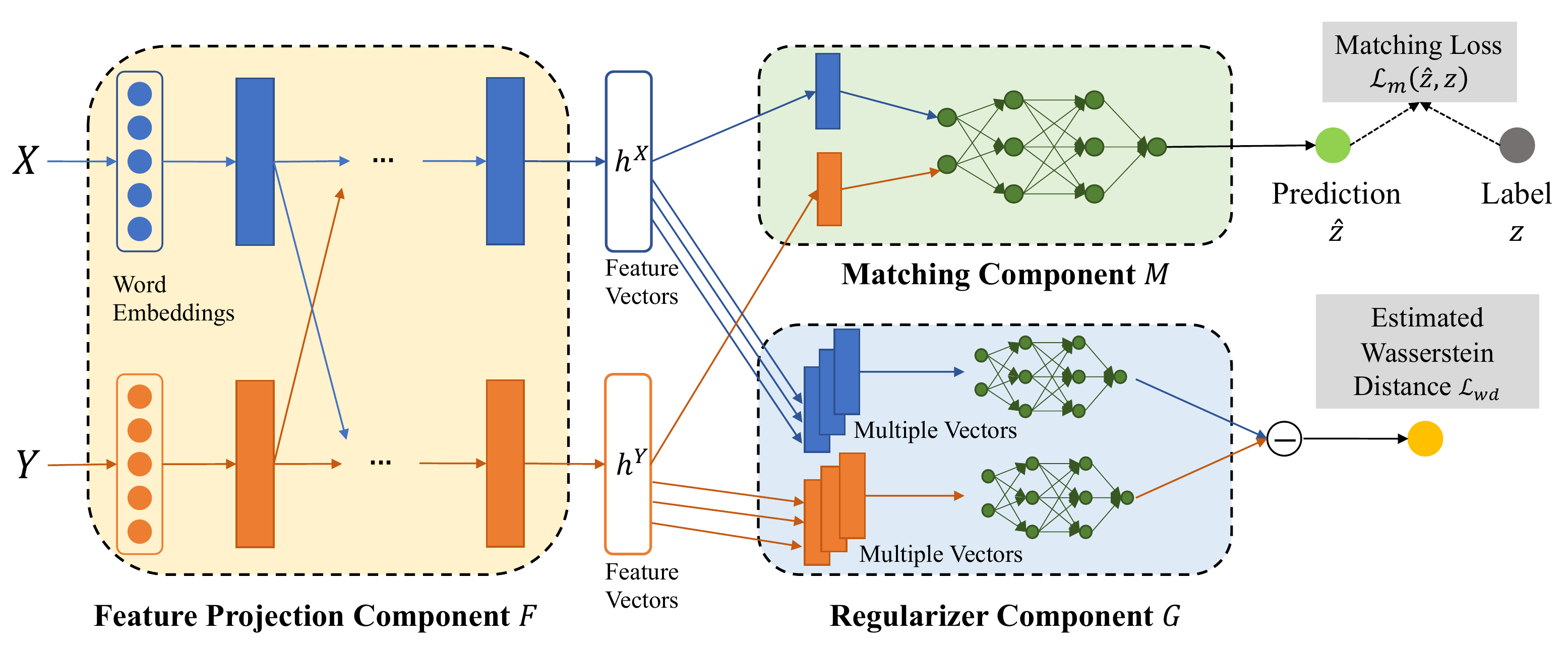}
\caption{WD-Match architecture. Note that the multiple parallel arrow lines to the regularizer component $G$ means that $G$ takes a set of feature vectors (based on a batch of sequence pairs), rather than one feature vector, as its inputs.
}\label{fig:WD-Match_arch}
\end{figure*}

{\bf The feature projection component}:  Given a sequence pair $(X,Y)$, it is first processed by the feature projection component $F$,
\[
[\mathbf{h}^X, \mathbf{h}^Y] = F(X, Y),
\]
where the feature projection function $F$ outputs a pair of $K$-dimensional feature vectors $\mathbf{h}^X, \mathbf{h}^Y$ in the semantic space. We suppose that $F$ is a neural network with a set of parameters $\mathbf{\theta}_F$ and all the parameters in $\mathbf{\theta}_F$ are sharing for $X$ and $Y$.

{\bf The matching component}: The output vectors from the feature projection component are then fed to the matching component $M$, 
\[
\hat{\mathbf{z}} = M([\mathbf{h}^X, \mathbf{h}^Y]),
\]
 $M$ outputs the predicted label $\hat{\mathbf{z}}$. We suppose that $M$ is a neural network with a set of parameters $\mathbf{\theta}_M$. 

{\bf The regularizer component:} Given two sets of the projected feature vectors $\mathbf{h}^X$ and $\mathbf{h}^Y$, the regularizer component estimates the Wasserstein distance between $\mathbb{P}_F^X$ and $\mathbb{P}_F^Y$, we denote $\mathbb{P}_F^X $ and $\mathbb{P}_F^Y$ are two distributions defined over the two groups of feature vectors $\mathbf{h}^X$ and $\mathbf{h}^Y$ respectively. 
\begin{equation}
\small
\begin{split}
\mathbb{P}^X_F \triangleq &P\left(\mathbf{h}^X| [\mathbf{h}^X, \mathbf{h}^Y] = F(X,Y) \wedge (X, Y)\sim  \mathcal{X}\times \mathcal{Y}\right),\\
\mathbb{P}^Y_F \triangleq &P\left(\mathbf{h}^Y| [\mathbf{h}^X, \mathbf{h}^Y] = F(X,Y) \wedge (X, Y) \sim \mathcal{X}\times \mathcal{Y}\right),
\end{split}
\end{equation}
where `$\sim$' means that the pairs $(X,Y)$ are sampled from the joint
space $\mathcal{X}\times \mathcal{Y}$.
Specifically, the Wasserstein distance between two probabilistic distributions $\mathbb{P}_{F}^X$ and $\mathbb{P}_{F}^Y$ is defined as:

\[
W(\mathbb{P}_F^X, \mathbb{P}_F^Y) = \inf_{\gamma\in\mathcal{J}(\mathbb{P}_F^X, \mathbb{P}_F^Y)}\int\|X - Y\|d\gamma(X,Y),
\]
where $\mathcal{J}(\mathbb{P}_F^X, \mathbb{P}_F^Y)$ denotes all joint distributions, $\gamma$ stands for $(X, Y)$ that have marginal distributions $\mathbb{P}_F^X$ and $\mathbb{P}_F^Y$. It can be shown that $W$ has the dual form~\cite{dual}:

\begin{equation}\label{eq:WDistance}
\begin{aligned}
W(\mathbb{P}_{F}^X, \mathbb{P}_{F}^Y) & =\sup_{|G|_L\le 1} {E}_{\mathbb{P}_{F}^X}[G(\mathbf{h}^X)]-{E}_{\mathbb{P}_{F}^Y}[G(\mathbf{h}^Y)],
\end{aligned}
\end{equation}
where `$|G|_L\le 1$' denotes that the `$\sup$' is taken over the set of all 1-Lipschitz\footnote{$G$ is 1-Lipschitz $\Leftrightarrow |G(\mathbf{h})-G(\mathbf{h}')|\leq |\mathbf{h}-\mathbf{h}'|$ for all $\mathbf{h}$ and $\mathbf{h}'$} function $G$; and function $G: \mathcal{R}^K\rightarrow \mathcal{R}$ maps each $K$-dimensional feature vector in the semantic space to a real number. In this paper, $G$ is set as a two-layer feed-forward neural network with a set of parameters $\mathbf{\mathbf}{\theta}_G$ clipped to $[-c,c]$, where $c>0 $ is a hyper-parameter.

Please note that different configurations of the feature projection component $F$, matching component $M$, and matching loss $\mathcal{L}_{m}$ leads to different matching models. Therefore, WD-Match can improve a matching model by setting the matching method as its underlying model. 

\subsection{Adversarial Training}
To learn the model parameters $\{\mathbf{\mathbf}{\theta}_F, \mathbf{\mathbf}{\theta}_M, \mathbf{\mathbf}{\theta}_G\}$, WD-Match sets up two training goals: minimizing the Wasserstein distance between $\mathbb{P}_{F}^X$ and $\mathbb{P}_{F}^Y$, and minimizing the loss in prediction in terms of the mistakenly predicted matching labels. The training process can be implemented under the adversarial learning framework and amounts to repeatedly interplays between two learning branches: the regularizer branch and the matching branch. 

In the regularizer branch, the objective term in the dual form Wasserstein distance (Equation~(\ref{eq:WDistance})) is approximately written as:
\[
\mathcal{O}_G(\theta_F, \theta_G) = \sum_{(X,Y) }\left[G(\mathbf{h}^X) - G(\mathbf{h}^Y)\right],
\]
where $[\mathbf{h}^X,\mathbf{h}^Y] = F(X, Y)$ are the projected feature vectors for $(X,Y)$. Maximizing $\mathcal{O}_G$ w.r.t. the parameters $\theta_G$ can achieve an approximation of the Wasserstein distance between $\mathbb{P}_{F}^X$ and $\mathbb{P}_{F}^Y$ in the semantic space defined by $F$:
\begin{equation}\label{eq:loss_wd}
    \mathcal{L}_{wd}(\mathbf{\theta}_F) = \max_{\mathbf{\theta}_G} \mathcal{O}_G (\theta_F, \theta_G).
\end{equation}
To make $G$ a Lipschitz function (up to a constant) and following the practices in~\cite{WGAN}, all of the parameters in $\theta_G$ are always clipped to a fixed range $[-c, c]$. In practice, the sequence pairs for training $G$ are randomly sampled from the training set $\mathcal{D}$. Note that $\mathcal{L}_{wd}$ still takes $\mathbf{\theta}_F$ as parameters because it is calculated on the basis of features generated by $F$.  

The matching branch simultaneously updates the matching network $M$ and feature projection network $F$ by seeking the minimization of the Wasserstein distance-regularized matching loss: 
\begin{equation}\label{eq:loss_adv}
\min_{\mathbf{\theta}_F, \mathbf{\theta}_M} \mathcal{L}_{reg} = \mathcal{L}_m (\mathbf{\theta}_F, \mathbf{\theta}_M) + \lambda \cdot \mathcal{L}_{wd} (\mathbf{\theta}_F),
\end{equation}
where $\lambda \in [0, 1]$ is a trade-off coefficient to balance the matching loss and regularizer, and $\mathcal{L}_m (\mathbf{\theta}_F, \mathbf{\theta}_M )$ is defined as
\[
\mathcal{L}_m (\mathbf{\theta}_F, \mathbf{\theta}_M ) = \sum_{(X, Y, z)\in\mathcal{D}} \ell_m(M(F(X, Y)), \mathbf{z}),
\]
where $\ell_m(\cdot,\cdot)$ is the matching loss function defined over each sequence-sequence-label triple in the training data. It can be, for example, the cross-entropy loss that measure the goodness of the predicted label $\hat{\mathbf{z}} = M(F(X, Y))$ by the matching network, compared to the ground truth label $\mathbf{z}$.

Algorithm~\ref{alg:WDMatch} shows the general procedure of WD-Match. WD-Match takes training set $\mathcal{D}={(X_i, Y_i, z_i)}_{i=1}^N$ and a number of hyper-parameters as inputs, and outputs the learned parameters $\theta_F$ and $\theta_M$. WD-Match run multiple rounds until convergence, and at each round it estimates the Wasserstein distance of the projected features and then update the projection component $F$ and matching component $M$. 
At each round, WD-Match alternatively maintains two branches. The regularizer branch updates the parameters $\mathbf{\theta}_G$, with the $\mathbf{\theta}_F$ fixed\footnote{Note that the regularizer does not depend on $M$, given $F$.}. It contains a sub-iteration in which the parameters are optimized in an iterative manner: First, objective $\mathcal{O}_G$ is constructed based on the sampled sequence pairs (line 4 - line 6); Then $\mathbf{\theta}_G$ is updated with gradient ascent (line 7); Finally, each parameter in $\mathbf{\theta}_G$ is clipped to $[-c,c]$ for satisfying the 1-Lipschitz constraint (line 8). The matching branch updates $\mathbf{\theta}_F$ and $\mathbf{\theta}_M$, with $\mathbf{\theta}_G$ fixed. It first samples another mini-batch data from the training data and estimates the regularized loss $\mathcal{L}_{adv}$ using the fixed $G$ (line 11 - line 13). Then, the gradients of the parameters is estimated and used to update the parameters (line 14).  

\begin{algorithm}[t] 
    \caption{The WD-Match algorithm.} 
    \label{alg:WDMatch} 
    \begin{algorithmic}[1] 
    \REQUIRE Training set $\mathcal{D}=\{(X_i, Y_i, \mathbf{z}_i)\}_{i=1}^N$; mini-batch sizes $n_1$ and $n_2$; adversarial training step $k$; trade-off coefficient $\lambda$; learning rates $\eta_1$ and $\eta_2$; clipping threshold $c$
    \REPEAT
    \STATE $\rhd$ Regularizer branch
    \FOR{$t=0$ to $k$}
    \STATE Sample a mini-batch $\{(X_i,Y_i,\mathbf{z}_i)\}^{n_1}_{i=1}$ from $\mathcal{D}$
    \STATE $[\mathbf{h}_i^X, \mathbf{h}_i^X] \leftarrow F(X_i, Y_i), \forall i=1,\cdots, n_1$
    \STATE $\mathcal{O}_G = \sum_{i=1}^{n_1}[G(\mathbf{h}_i^X) - G(\mathbf{h}_i^Y)]$
    \STATE $\mathbf{\theta}_G \leftarrow \mathbf{\theta}_G +\eta_1 \bigtriangledown_{\mathbf{\theta}_G} \mathcal{O}_G$ \COMMENT{Eq.~(\ref{eq:loss_wd})}
    \STATE ClipWeights$(\theta_G, -c, c)$
    \ENDFOR
    \STATE $\rhd$ Matching branch
    \STATE Sample a mini-batch $\{(X_i,Y_i,z_i)\}^{n_2}_{i=1}$ from $\mathcal{D}$
    \STATE $[\mathbf{h}_i^X, \mathbf{h}_i^Y] \leftarrow F(X_i, Y_i), \forall i=1,\cdots, n_2$
    \STATE $\mathcal{L}_{reg} = \sum_{i=1}^{n_2}[\ell_m(M(F(X,Y)), \mathbf{z}_i) + \lambda [G(\mathbf{h}_i^X) - G(\mathbf{h}_i^Y)]]$
    \STATE $\{\mathbf{\theta}_F, \mathbf{\theta}_M\}\leftarrow \{\mathbf{\theta}_F, \mathbf{\theta}_M\} - \eta_2  \bigtriangledown_{\mathbf{\theta}_F,\mathbf{\theta}_M } \mathcal{L}_{reg}$ \COMMENT{Eq.~(\ref{eq:loss_adv})}
    \UNTIL{convergence} 
    \RETURN $\{\mathbf{\theta}_F,\mathbf{\theta}_M\}$
    \end{algorithmic}
    \end{algorithm}

\section{Experiments}
We conducted experiments to test the performances of WD-Match, and analyzed the results.

\subsection{Datasets and Metrics}
We use four large scale publicly matching benchmarks: SNLI (Stanford Natural Langauge Inference) ~\cite{SNLI}, SciTail ~\cite{Scitail}, TrecQA ~\cite{TRECQA}, WikiQA ~\cite{wikiqa}. Table~\ref{tab:datasets} provides a summary of the datasets used in our experiments.

\begin{table}[t]
    \centering
    \caption{Statistics of four dataset used in our experiment, $|C|$ denotes the number of classes and R denotes a ranking formulation.}
    \label{tab:datasets}
    \begin{tabular}{llcc}
    \hline
    \textbf{Dataset} & \textbf{Task} & \textbf{$|C|$} &\textbf{Pairs}\\ 
    \hline
    \hline
    SNLI&premise-hypothesis&3&570k\\
    SciTail&premise-hypothesis&2&27k\\
    TrecQA&question-answer&R&56k\\
    WikiQA&question-answer&R&20k\\
    \hline
    \end{tabular}
    
\end{table}

\textbf{SNLI}~\footnote{\url{https://nlp.stanford.edu/projects/snli}} is a benchmark for natural language inference. In SNLI, each data record is a premise-hypothesis-label triple. The premise and hypothesis are two sentences and the label could be ``entailment'', ``neutral'', ``contradiction'', or ``-''. In our experiments, following the practices in~\cite{SNLI}, the data with label ``-'' are ignored. We follow the original dataset partition. Accuracy is used as the evaluation metric for this dataset.

\textbf{SciTail}~\footnote{\url{http://data.allenai.org/scitail/}} is an entailment dataset based on multiple-choice science exams and web sentences. Each record is a premise-hypothesis-label triple. The label is ``entailment'' or ``neutral'', because scientific factors cannot contradict. We follow the original dataset partition. Accuracy are used as the evaluation metric for this dataset.

\textbf{TrecQA}~\footnote{\url{https://github.com/castorini/NCE-CNN-Torch/tree/master/ data/TrecQA}} is a answer sentence selection dataset designed for the open-domain question answering setting. We use the raw version TrecQA, questions with no answers or with only positive/negative answers are included. The raw version has 82 questions in the development set and 100 questions in the test set. Mean average precision (MAP) and mean reciprocal rank (MRR) are used as the evaluation metrics for this task. 

\textbf{WikiQA}~\footnote{\url{https://www.microsoft.com/en-us/download/details.aspx?id=52419}} is a retrieval-based question answering dataset based on Wikipedia. We follow the data split of original paper. This dataset consists of 20.4k training pairs, 2.7k development pairs, and 6.2k testing pairs. We use MAP and MRR as the evaluation metrics for this task. 

\subsection{Experimental Setup}
\label{sec:model}

In WD-Match, different configurations of the feature projection component $F$ and matching component $M$ lead to different matching models. In the experiments,
RE2~\cite{RE2}, DecATT~\cite{DecAtt}, CAFE~\cite{CAFE} and BERT~\cite{BERT} were set as the underlying models, achieving new models respectively denoted as ``WD-Match (RE2)'', ``WD-Match (DecAtt)'', ``WD-Match (CAFE)'', and  ``WD-Match (BERT)''.

Specifically, in WD-Match(RE2), $F$ is a stacked blocks which consist of multiple convolution layers and multiple attention layers, and $M$ is an MLP; in WD-Match(DecAtt), $F$ is an attention layer and a aggregation layer, $M$ is an MLP. Please note that we did not implement the Intra-Sentence Attention in our experiments; in WD-Match(CAFE), $F$ is a highway encoder with a alignment layer and a factorization layer and $M$ is another highway network. Please note that we remove the character embedding and position embedding in our experiments; in WD-Match(BERT), $F$ is a pre-trained BERT-base\footnote{\url{https://github.com/google-research/bert}} model, $M$ is an MLP. Please note that for easing of combining with WD-Match, BERT was only used to extract the sentence features separately in our experiments. 
The $G$ module for four models are identical: a non-linear projection layer and a linear projection layer.

For all models, the parameters of $F$ and $M$ were directly set as its original settings. In the training, all models were trained using the Adam optimizer with learning rate $\eta_2$ tuned amongst \{0.0001, 0.0005, 0.001\}. Batch
size $n_2$ was tuned amongst \{256, 512, 1024\}. The trade-off coefficient $\lambda$ was tuned from [0.0001, 0.01]. Clipping threshold was tuned from [0.1, 0.5].
Word embeddings were initialized
with GloVe~\cite{glove} and fixed during training. We implemented WD-Match models in Tensorflow.

\subsection{Experimental Results}
\label{expres}

Table~\ref{tab:Exp:SNLI} reports the results of WD-Match and the popular baselines on the SNLI test set. The baselines results are reported from their original papers. From the results, we found that WD-Match (RE2) outperformed all of the baselines, including the underlying model RE2. The results indicate the effectiveness of WD-Match and its Wasserstein distance-based regularizer in the asymmetric matching tasks of natural language inference. We further tested the performances of WD-Match (DecAtt) and WD-Match(BERT) which used DecAtt and BERT as the underlying matching models, respectively, to show whehter WD-Match can improve a matching method by using the method as its underlying model. From the results shown in Table~\ref{tab:Exp:SNLI}, we can see that on SNLI, WD-Match (DecAtt) ourperform DecAtt in terms of accuracy. Similarly, WD-Match (BERT) improved BERT about 0.4 points in terms of accuracy.
\begin{table}[t]
    \caption{Performance comparison on SNLI test set.}\label{tab:Exp:SNLI}
    \centering
    \begin{tabular}{@{ }l@{ }cc@{ }}
        \hline
         Models  & Acc.(\%) &\#Params\\
        \hline 
        \hline
        BiMPM~\cite{BiMPM} & 86.9 & 1.6M\\
        ESIM~\cite{ESIM/HIM} & 88.0& 4.3M\\
        DIIN~\cite{DIIN} & 88.0 & 4.4M\\
        MwAN~\cite{MwAN} & 88.3 &14M\\
        HIM~\cite{ESIM/HIM} & 88.6 & 7.7M\\
        SAN ~\cite{SAN} & 88.6 & 3.5M\\
        CSRAN~\cite{CSRAN"} &  88.7& 13.9M\\
        DRCN ~\cite{DRCN}& 88.9 & 6.7M\\
        \hline
        RE2 ~\cite{RE2} & 89.0 &2.8M\\
        \textbf{WD-Match (RE2)} &\textbf{89.1} &2.9M\\    
        \hline
        DecAtt~\cite{DecAtt}& 82.5& 0.26M \\
        \textbf{WD-Match (DecAtt)}& \textbf{82.6} &0.30M\\
        \hline
        BERT~\cite{BERT}& 83.7 & 0.11B\\
        \textbf{WD-Match (BERT)}& \textbf{84.1} & 0.11B\\
        \hline
    \end{tabular}
\end{table}

Table~\ref{tab:Exp:SciTail} reports the results of WD-Match and the baselines on the SciTail test set. The baselines results are reported from the original papers. We found that WD-Match (RE2) outperformed all of the baselines. The result further confirm WD-match's effectiveness in the asymmetric matching task of scientific entailment. We also tested the performances of WD-Match (DecAtt) and WD-Match(BERT) on Scitail dataset. From the results shown in Table~\ref{tab:Exp:SciTail}, we can see that WD-Match (DecAtt) improved DecAtt 1.2 points in terms of accuracy. Similarly, WD-Match (BERT) improved BERT about 2.7 points in terms of accuracy. The results verified that WD-Match's ability in improving its underlying model.

\begin{table}[t]
    \caption{Performance comparison on SciTail test set}    \label{tab:Exp:SciTail}
    \centering
    \begin{tabular}{lc}
        \hline
         Models  & Acc.(\%)\\
        \hline \hline
        ESIM~\cite{ESIM/HIM} & 70.6\\
        DGEM~\cite{Scitail} & 77.3\\
        HCRN~\cite{HCRN} & 80.0\\
        CSRAN~\cite{CSRAN"} &  86.7\\
        \hline
        RE2~\cite{RE2} & 86.6 \\
        \textbf{WD-Match (RE2)} &
        \textbf{87.0}\\    
        \hline
        BERT~\cite{BERT}& 79.2 \\
        \textbf{WD-Match (BERT)}& \textbf{81.9} \\
        \hline
        DecAtt~\cite{DecAtt}& 81.7 \\
        \textbf{WD-Match (DecAtt)}& \textbf{82.9} \\
        \hline
    \end{tabular}
\end{table}

Table~\ref{tab:Exp:wiki} reports the results of WD-Match and the baselines on the WikiQA test set. The baselines result are reported from the original papers. Following RE2, point-wise binary classification loss rather than pairwise ranking loss was used to train the model. The best hyperparameters including early stopping were tuned on WikiQA development set. From the results we can see that WD-Match (RE2) obtained a better result in terms of MAP and MRR on  WikiQA. 
To further verify the effectiveness of WD-Match on QA task, we incorporated DecAtt and CAFE~\cite{CAFE} into WD-Match, then compare their performance to the respective underlying models on WikiQA and TrecQA datasets. Table~\ref{tab:Exp:wiki} and Table~\ref{tab:trecqamodel} report the experimental results on WikiQA test set and TrecQA test set respectively. Similarly, WD-Match outperformed its underlying model on both datasets.


\begin{table}[t]
    \caption{Performance comparison on WikiQA test set.}\label{tab:Exp:wiki}
    \centering
    \begin{tabular}{@{ }l@{  }c@{  }c@{ }}
        \hline
         Models  & MAP(\%) &MRR(\%)\\
        \hline \hline
        KVMN~\cite{KVMN} & 70.69& 72.65\\
        BiMPM~\cite{BiMPM} & 71.80& 73.10\\
        IWAN~\cite{IWAN} & 73.30& 75.00\\
        CA~\cite{CA} & 74.33& 75.45\\
        HCRN~\cite{HCRN}& 74.30& 75.60\\
        \hline

        RE2~\cite{RE2} & 74.96& 76.58 \\
       \textbf{WD-Match (RE2)}&\textbf{75.31}&\textbf{76.89}\\   
       \hline
        DecAtt~\cite{DecAtt}& 64.03 & 65.92 \\
        \textbf{WD-Match (DecAtt)}& \textbf{65.16} & \textbf{67.24} \\
        \hline
        CAFE~\cite{CAFE}& 64.19 & 65.65\\
        \textbf{WD-Match (CAFE)}& \textbf{66.36} & \textbf{67.59}\\
        \hline
    \end{tabular}
\end{table}

\begin{table}[t]
    \centering
    \caption{Performance comparison on TrecQA test set.}
    \label{tab:trecqamodel}
    \begin{tabular}{@{ }l@{  }c@{  }c@{ }}
    \hline
    Model & MAP(\%) & MRR(\%) \\ 
    \hline
    \hline
    DecAtt~\cite{DecAtt}& 70.62 & 76.88 \\
    \textbf{WD-Match (DecAtt)}& \textbf{72.30} & \textbf{76.91}\\
    \hline
    CAFE~\cite{CAFE}& 65.00 & 71.86\\
    \textbf{WD-Match (CAFE)}& \textbf{67.49} & \textbf{73.05}\\
    \hline
    \end{tabular}
\end{table}

We list the number of parameters of different text matching models in Table~\ref{tab:Exp:SNLI}. Compared to the underlying model, the additional parameters of WD-Match come from the regularizer component $G$. We can see that the parameters of the regularizer component $G$ are far less than the underlying model. 
$G$ module is implemented as a two-layer MLP (the number of neurons in the second layer is set as one). Therefore, the additional computing cost comes from the training of the two-layer MLP, which is of $O(T*N*K*1)$, where $T$ is the number of training iterations, $N$ number of training examples, $K$ number of neurons in the first layer of MLP (without considering the compute cost of the activation function). We can see that the additional computing overhead is much lower than that of the underlying methods which usually learn much more complex neural networks for the feature projection and the matching. \\

Summarizing the results above and the results reported in Section~\ref{expres}, we can conclude that WD-Match is a general while strong framework that can improve different matching models by using them as its underlying matching model.

\subsection{Visualization of the Distributions of Feature Vectors}
\begin{figure}[t]
     \centering
    \includegraphics[width=0.5\textwidth]{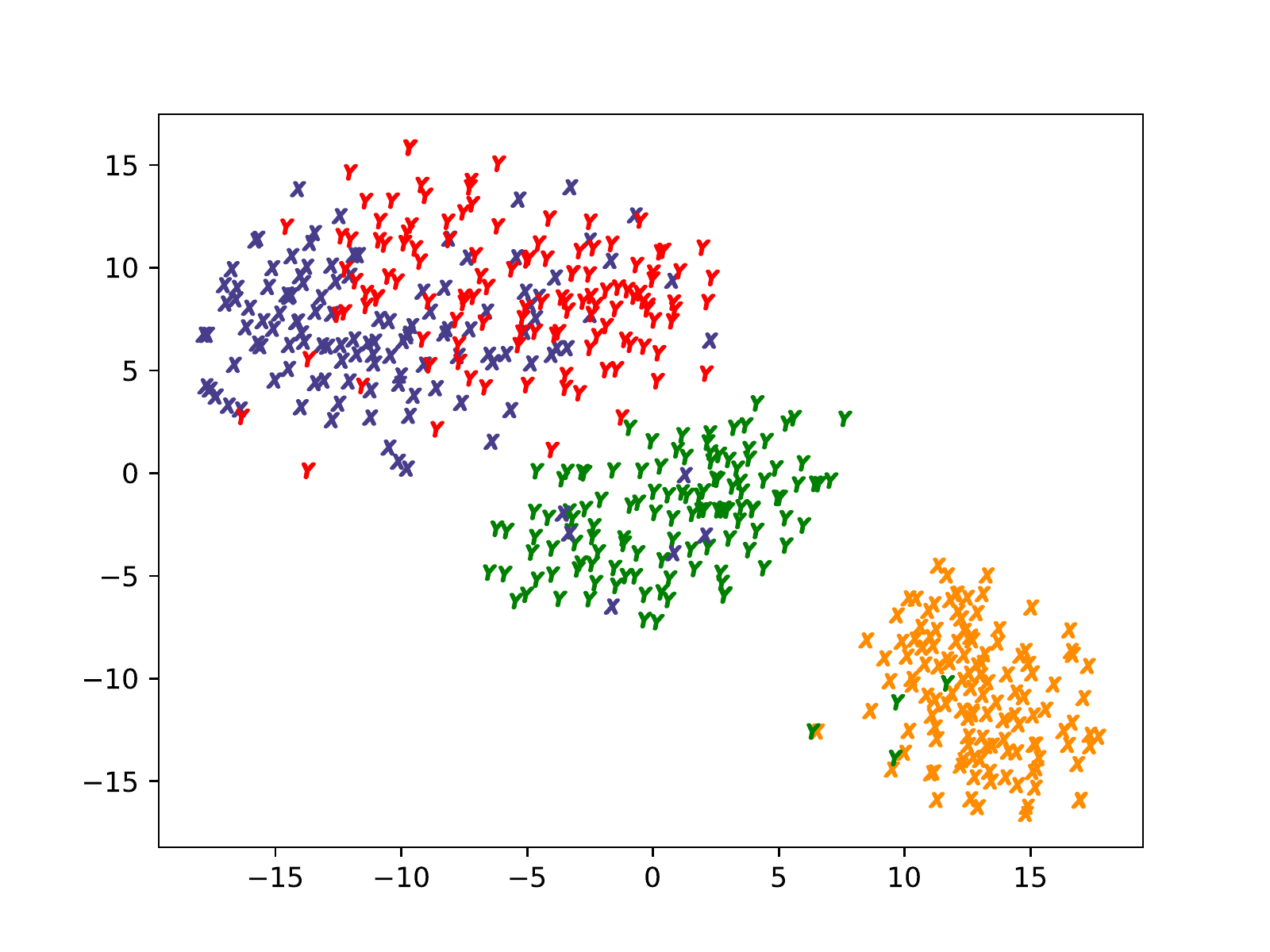}
    \caption{t-SNE visualization of the projected feature vectors, based on the RE2 and WD-Match (RE2) models trained on SciTail dataset. This figure illustrates the vector distributions after 5 training steps. The orange `$X$' and green `$Y$' correspond to $\mathbb{P}^X_F$ and $\mathbb{P}^Y_F$ of RE2, The dark blue  `$X$' and red `$Y$' correspond to $\mathbb{P}^X_F$ and $\mathbb{P}^Y_F$ of WD-Match (RE2), respectively.}
    \label{fig:t-SNE}
\end{figure}

Figure~\ref{fig:iter}(a) shows that there exists a gap between two feature vectors, due to the heterogenous nature of the texts from two asymmetrical domains. 
We conducted experiments to analyze how the feature vectors (i.e., $\mathbf{h}^X$ and $\mathbf{h}^Y$) generated by WD-Match distributed in the common semantic space, using WD-Match(RE2) as an example.
Specifically, we trained a RE2 model and a WD-Match (RE2) model based on SciTail dataset. Note that in this experiment, the adversarial training step $k$ is set as 5, that is, WD-Match (RE2) repeats regularizer branch for 5 times before matching branch.
We recorded all of the training feature vectors (i.e., $\mathbf{h}^X$
and $\mathbf{h}^Y$) and illustrated them in the Figure~\ref{fig:t-SNE} by t-SNE . The orange `$X$' and green `$Y$' correspond to $\mathbb{P}^X_F$ and $\mathbb{P}^Y_F$ of RE2, The dark blue  `$X$' and red `$Y$' correspond to $\mathbb{P}^X_F$ and $\mathbb{P}^Y_F$ of WD-Match (RE2), respectively. As we can see from Figure~\ref{fig:t-SNE}, the feature vectors from RE2 are separately distributed while the feature vectors from WD-Match (RE2) are indistinguishable. 
It demonstrates that compared to the underlying model RE2, WD-Match (RE2) distributes the feature vectors in semantic space better and faster. 

\subsection{Convergence and Effects of Wasserstein Distance-based Regularizer}
We conducted experiments to test 
how the Wasserstein distance-based regularizer guides the training of matching models. 
\begin{figure}[t]
     \centering
    \includegraphics[width=0.5\textwidth]{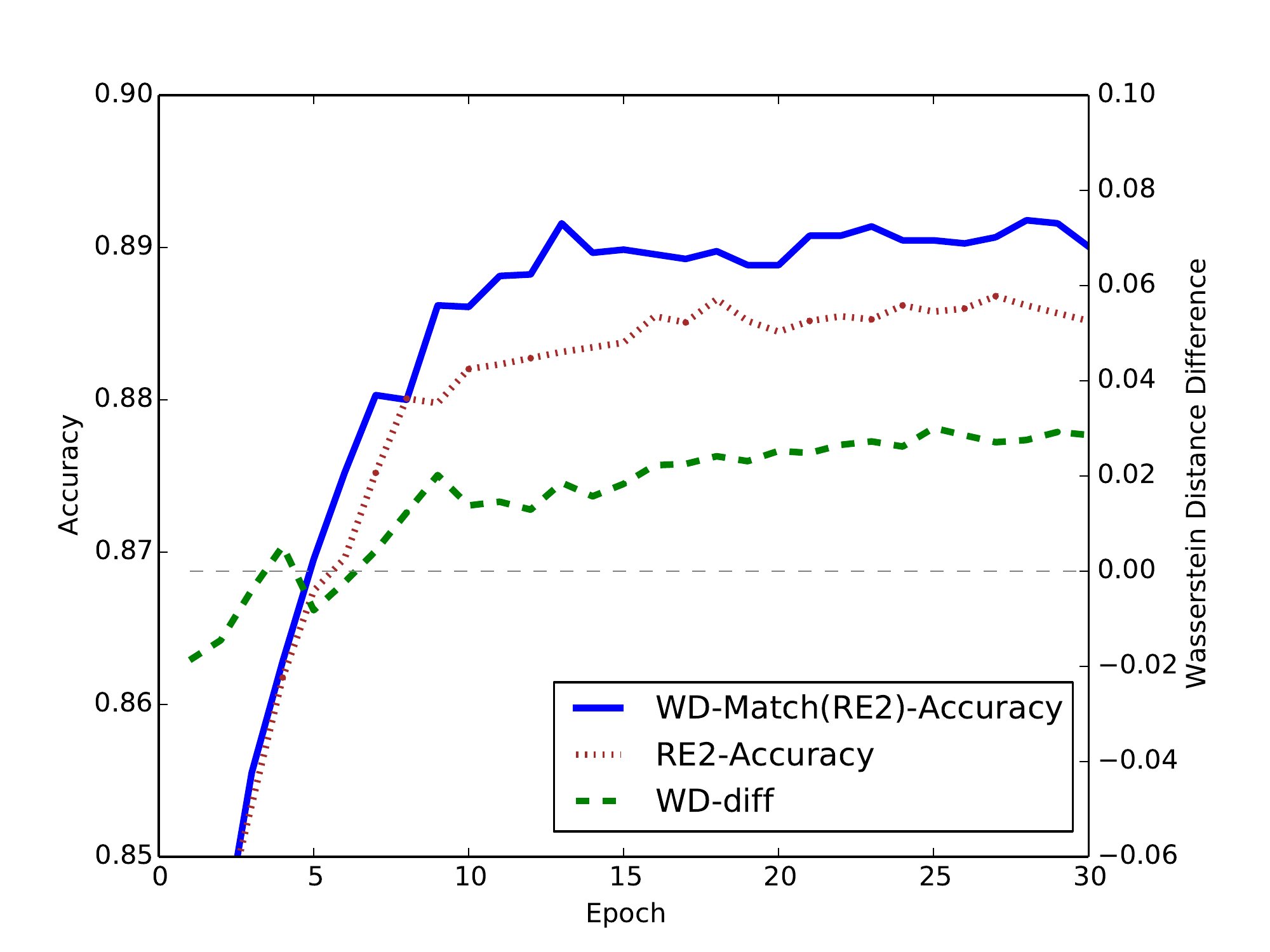}
    \caption{Accuracy curves and Wasserstein distance difference curve w.r.t. training epochs for RE2 and \mbox{WD-Match (RE2)}.}
    \label{fig:trainCurve}
\end{figure}

Specifically, we tested the WD-Match (RE2) and RE2 models generated at each training epochs. The accuracy curve on the basis of development set of SNLI was illustrated in Figure~\ref{fig:trainCurve} (denoted as ``WD-Match (RE2)-Accuracy'' and ``RE2-Accuracy''). Comparing these two training curves, we can see that WD-Match (RE2) outperformed RE2 when the training closing to converge (after about 15 epochs). We can conclude that WD-Match (RE2) obtained higher accuracy than RE2.

To investigate how the Wasserstein distance guides the training of matching models, we recorded the estimated Wasserstein distances at all of the training epochs of RE2 and WD-Match (RE2) based on the converged $G$ network. The curve ``WD-Diff'' shows the differences between the Wasserstein distances by RE2 and that of by WD-Match (RE2) at each of the training epoch (i.e., $\mathcal{L}_{wd}(\theta_F)$ of RE2 minus $\mathcal{L}_{wd}(\theta_F)$ of WD-Match (RE2)).
From the curve we can see that at the beginning of the training (i.e., epoch 1 to 5), the ``WD-Diff'' was near to zero. With the training went on (i.e., epoch 5 to 30), the Wasserstein distance by WD-Math(RE2) became smaller than that of by RE2 (the WD-Diff curve is above the zero line), which means that WD-Match (RE2)'s feature projection module $F$ was guided to move feature vectors together more thoroughly and faster, which are more suitable for matching. The results indicate WD-Match achieved its design goal of guiding the distributions of the projected feature vectors. 


It is interesting to note that, comparing all of the three curves in Figure~\ref{fig:trainCurve}, we found the WD-Diff curve is close to zero at the beginning of the training, and the accuracy curves of WD-Match (RE2)-Accuracy and RE2-Accuracy are similar at the beginning. With the training went on (after epoch 10), the Wasserstein distance differences became larger. At the same time, the accuracy gaps (between WD-Match (RE2)-Accuracy and RE2-Accuracy) also become larger. The results clearly reflect the effects of Wasserstein distance-based regularizer: minimizing the regularizer leads to better distribution of feature vectors in terms of matching. 

\section{Conclusion and Future Work}
In this paper, we proposed a novel Wasserstein distance-based regularizer to improve the sequence representations, for text matching in asymmetrical domains. The method, called WD-Match, amounts to adversarial interplay of two branches: estimating the Wasserstein distance given the projected features, and minimizing the Wasserstein distance regularized matching loss. We show that the regularizer helps WD-Match to well distribute the generated feature vectors in the semantic space, and therefore more suitable for matching. Experimental results on four benchmarks showed that WD-Match can outperform the baselines including its underlying models. Empirical analysis showed the effectiveness of the Wasserstein distance-based regularizer in text matching. 

In the future, we plan to study different regularizers in the asymmetrical text matching task, for further exploring their effectiveness in bridging the gap between asymmetrical domains.

\section*{Acknowledgments}
This work was funded by the National Key R\&D Program of China (2019YFE0198200), the National Natural Science Foundation of China (61872338, 61832017, 61906180, 62006234), Beijing Academy of Artificial Intelligence (BAAI2019ZD0305), and Beijing Outstanding Young Scientist Program (BJJWZYJH012019100020098).

\bibliography{anthology,emnlp2020}
\bibliographystyle{acl_natbib}



\end{document}